\pdfoutput=1
\documentclass[conference]{IEEEtran}
\usepackage[letterpaper,left=56.5bp,right=56.5bp,top=55bp,textheight=680pt,columnsep=12pt]{geometry}

\usepackage[hidelinks]{hyperref}

\IEEEoverridecommandlockouts

\usepackage{cite}
\usepackage{amsmath,amssymb,amsfonts}
\usepackage{algorithmic}
\usepackage{graphicx}
\usepackage{textcomp}
\usepackage{xcolor}
\usepackage{multirow}
\usepackage{float}
\usepackage{siunitx}
\usepackage{tikz}
\usetikzlibrary{arrows.meta,calc,positioning,shapes.geometric}
\usepackage{algorithm}
\usepackage{bm}
\usepackage{subcaption}
\usepackage{booktabs}

\usepackage{pifont}

\newcommand{\R}{\mathbb{R}}

\def\BibTeX{{\rm B\kern-.05em{\sc i\kern-.025em b}\kern-.08em
    T\kern-.1667em\lower.7ex\hbox{E}\kern-.125emX}}
\begin{document}

\title{Steerable and Reactive Grasping Through Modular Design with a Three-Point Interface}

\author{
    Andrew Nguyen$^{1,2}$, Yonghyeon Lee$^{3, \dagger}$, Sangbae Kim$^{2, \dagger}$\\
    \thanks{$^1$University of Michigan, Ann Arbor, MI, USA. $^2$Department of Mechanical Engineering, MIT, Cambridge, MA, USA. $^3$Department of AI, Yonsei University, Seoul, South Korea. Emails: \texttt{andyng@umich.edu, andyng@mit.edu, sangbae@mit.edu, yonghyeon.lee@yonsei.ac.kr}. $^\dagger$ Corresponding authors.}
    }

\maketitle

\begin{abstract}
Dexterous grasping requires deciding where to grasp, reaching the target, and maintaining stable contact. We connect these stages through a compact three-point interface that separates global geometric reasoning from local contact control. Given object geometry and optional language commands, our framework samples contact triples from a precomputed grasp-affordance heatmap. A model-based reactive controller tracks the object, avoids collisions, and guides the hand toward the selected contacts. In the final centimeters, a Reinforcement Learning (RL) policy uses proprioceptive feedback to refine and stabilize the grasp despite reaching and perception errors. It observes only finger joint states and its recent actions, with no target points, visual observations, or object geometry, so a single policy is shared across objects and grasp configurations. In simulation, we compare grasp-and-lift success against squeeze and end-to-end baselines, characterize reaching convergence, and demonstrate grasp steering; hardware demonstrations on two training objects and one unseen object illustrate the full pipeline. Our modular framework uses geometry to guide the reach and local feedback to secure the grasp. \textbf{Project Website:} \href{https://heatmap-three-point-interface-grasping.github.io/}{https://heatmap-three-point-interface-grasping.github.io/}

\end{abstract}


\section{Introduction}
\label{sec:introduction}

Robotic manipulation begins with grasping~\cite{bohg2014data}. Grasping typically involves three stages: selecting where to grasp~\cite{ferrari1992planning}, moving the arm and hand toward the selected region while avoiding collisions~\cite{berenson2011task, LaValle_2006}, and regulating contact forces during the final centimeters of interaction to establish a stable grasp~\cite{ferrari1992planning, nguyen1988constructing, li2003computing}. Ideally, these stages should operate reactively, continuously adapting to feedback rather than executing a fixed sequence. Achieving this behavior requires real-time replanning and control across both free-space motion and contact-rich interaction~\cite{kappler2018real, kroemer2010combinig, lee2026hierarchical}.

Recent work on hierarchical reactive grasping addresses the reaching component of this challenge~\cite{lee2026hierarchical}. Given predefined fingertip targets for two-finger pinch grasps, it provides a reactive, collision-free motion planning and control framework that guides the hand toward a candidate grasp. Its scope, however, is primarily reactive reaching with two fingers. Under a point-contact model without torsional friction, two-finger pinch grasps cannot provide full force closure because rotation about the line connecting the contacts remains unconstrained. Their final force-control problem is nevertheless comparatively simple, primarily requiring balanced squeezing forces~\cite{nguyen1988constructing, li2003computing}. Furthermore, grasp targets are predefined for known objects, leaving target generation outside the framework. Extending this approach therefore requires both a method for selecting suitable contacts and a controller that can stabilize more complex contact configurations.

Building on the formulation in~\cite{lee2026hierarchical}, we propose a steerable three-finger pinch-grasping framework that integrates target selection, reactive reaching, and grasp stabilization. The target grasp region can be adjusted in real time according to high-level context, such as a language instruction, while three suitably placed contacts can establish force closure~\cite{nguyen1988constructing, li2003computing}. This extension presents two challenges. First, the system must select three surface points that support a stable grasp while satisfying the requested context. Second, the stabilization controller must accommodate residual reaching and perception errors. Reaching may terminate before the fingertips align precisely with their targets, and uncertainty in the estimated object pose can further alter the resulting contacts. Reliable grasping therefore requires a feedback policy that refines the grasp during contact, rather than simply closing the fingers around the estimated object location.

\begin{figure*}[!t]
    \centering
    \includegraphics[width=1.0\linewidth]{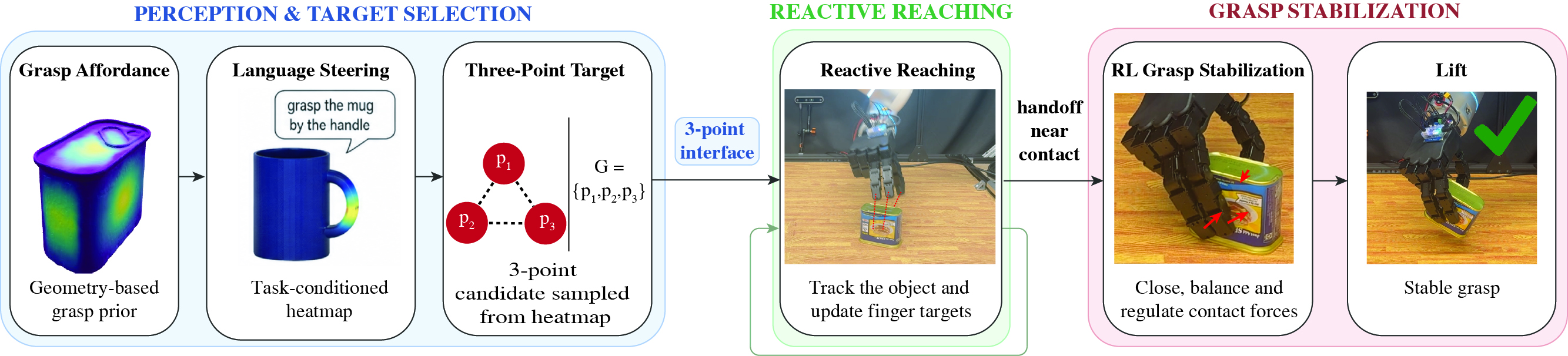}
    \caption{Our system consists of three stages. (Perception \& Target Selection) A task-agnostic grasp-affordance heatmap is computed from the object's geometry using a geometry-based grasp prior, optionally refined by a language instruction into a task-conditioned heatmap; a three-point grasp candidate is sampled from it. (Reactive Reaching) The three points are passed through a three-point interface to a model-based reactive controller, which tracks the object and updates finger targets until near contact. (Grasp Stabilization) A learned policy then closes the fingers, refines contact, stabilizes the grasp, and lifts the object.}
    \label{fig:grasping-pipeline}
\end{figure*}

Our framework addresses these challenges through the three-stage pipeline shown in Fig.~\ref{fig:grasping-pipeline}. We generate geometry-based grasp-affordance heatmaps offline and use them to sample three-point grasp candidates in real time, with optional language conditioning to steer contact selection toward a requested region. The reactive reaching method of~\cite{lee2026hierarchical} then guides the fingers toward the selected contacts. Finally, a reinforcement learning policy uses proprioceptive feedback to refine and stabilize the grasp during the final centimeters of interaction.

In contrast to end-to-end or geometry-conditioned reinforcement learning approaches to dexterous grasping~\cite{qin2023dexpoint,liu2023dexrepnet, lum2024dextrah,zhang2025robustdexgrasp}, our framework explicitly separates geometric reaching from contact-rich grasp refinement. In end-to-end formulations, a single learned policy often handles object geometry, robot configuration, global reaching motion, and local contact interactions. Variations in object pose, approach direction, support height, and surrounding geometry can therefore place additional demands on the policy's training distribution. For example, a policy trained primarily on top-down grasps at a fixed table height may need to generalize beyond its training conditions to execute a side grasp or operate at a different support height, even when the desired contact configuration remains geometrically similar.

A compact interface of three target contact points allows our framework to assign these responsibilities to separate components. The model-based reactive planner handles global reaching and adapts to changes in object pose and environment geometry, subject to grasp reachability and collision constraints. The learned policy handles only local grasp refinement and stabilization. It receives only proprioceptive observations: the three target contact points set its pre-grasp pose and shape its training reward, but neither they nor the object geometry nor raw visual input are observed. This division reduces the scope of the learned control problem and its direct dependence on object-specific representations, so the same policy is reused across objects and grasp configurations without retraining.

Experiments include a reach-convergence sweep, which reveals residual errors in model-based reaching, and a closing-controller comparison, which separates where the two controllers fail: naive squeezing slips the object after lift-off roughly ten times as often. Our modular system achieves higher grasp-and-lift success rates than naive squeezing and end-to-end RL baselines. We further evaluate anchor-based grasp steering and demonstrate the system on hardware using two training objects and one unseen object.

Our contributions are summarized as follows:
\begin{itemize}\setlength{\itemsep}{0pt}
\item a three-point interface that generates contact targets rather than requiring them as input, combining an offline grasp-affordance heatmap with online selection of three contacts adapted to the current hand pose;
\item anchor-based steering that re-centers the heatmap toward a specified region within milliseconds, without retraining or modifying downstream modules;
\item a last-centimeter RL policy for three-finger contact acquisition and lifting that, using only finger joint states and recent actions, secures grasps that naive squeezing drops.
\end{itemize}

\section{Methods}
\label{sec:method}

\subsection{Steerable Heatmap}
\label{sec:steerable-heatmap}

\begin{figure}[!t]
    \vspace{-5pt}
    \centering
    \includegraphics[width=1.0\linewidth]{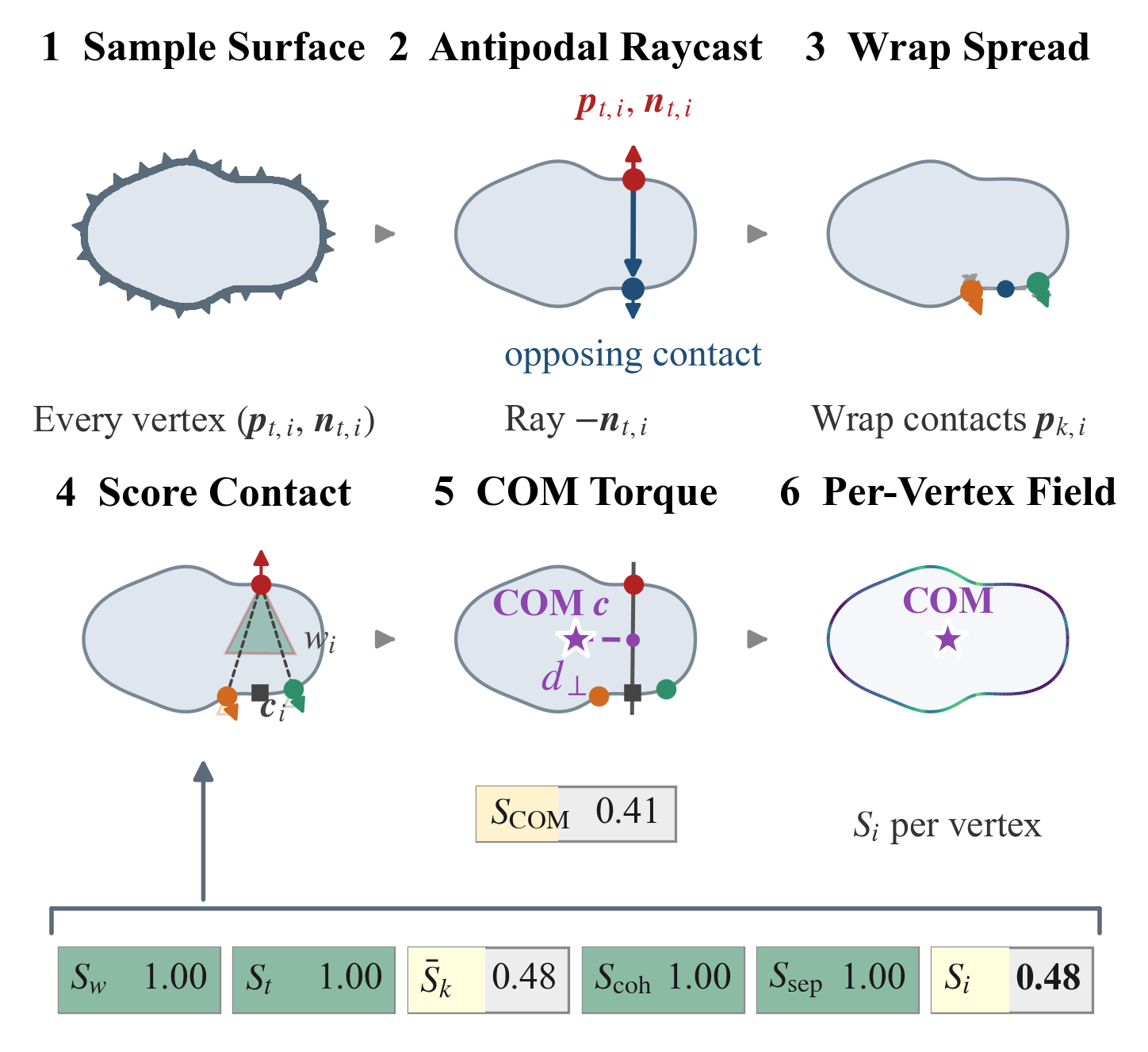}
    \caption{\textbf{Offline heatmap generation algorithm:} (1) select a thumb vertex and normal; (2) cast an inward ray to the opposing surface; (3) place two wrap contacts; (4) score aperture, friction-cone alignment, normal coherence, and finger separation; (5) weight by grasp-axis distance; and (6) repeat over all vertices to obtain the heatmap.}
    \label{fig:heatmap-generation-pipeline}
    \vspace{-10pt}
\end{figure}

Building on antipodal and analytic grasp-quality formulations~\cite{nguyen1988constructing,ferrari1992planning,mahler2017dexnet}, we precompute a grasp-affordance heatmap for each object by exhaustively evaluating antipodal contact configurations over its mesh vertices. Fig.~\ref{fig:heatmap-generation-pipeline} illustrates the procedure. Each vertex $i$ defines a candidate thumb contact $(\bm p_{t,i}, \bm n_{t,i})$, where $\bm n_{t,i}$ is the outward surface normal. We cast an inward ray along $-\bm n_{t,i}$ and use its farthest intersection with the mesh as the opposing contact.
For our three-finger hand, we initialize two wrap contacts at evenly spaced offsets around this contact, then project them onto the nearest points on the mesh.
Wrap contacts are distributed along the local long axis of the surface—the tangent direction in which the surface normal changes least. Thus, on curved objects such as cylinders, the contacts spread lengthwise rather than around the circumference, keeping their normals antipodal to the thumb.

Let $\bm c_i$ denote the centroid of the wrap contacts, $w_i = \|\bm c_i-\bm p_{t,i}\|$ the effective aperture, and $\bm g_i=(\bm c_i-\bm p_{t,i})/w_i$ the grasp axis. We compute the geometric affordance as
\begin{equation}
    S_i = S_w(i)\, S_t(i)\, \left(\frac{1}{2}\sum_{k=1}^{2}S_k(i)\right) S_{\mathrm{coh}}(i)\, S_{\mathrm{COM}}(i)\, S_{\mathrm{sep}}(i).
\label{eq:affordance}
\end{equation}
Here, $S_w$ penalizes grasp widths outside the gripper’s feasible span. $S_t$ and $S_k$ penalize angular misalignment between the grasp axis and the inward-facing normals at the thumb and wrap contacts ($k=1,2$), respectively; each is a Gaussian in the misalignment angle scaled by the friction-cone half-angle $\arctan\mu$, where $\mu$ is the friction coefficient, so contacts outside the friction cone score near zero. $S_{\mathrm{coh}}$ penalizes wrap contacts on surfaces facing different directions, such as contacts that straddle a sharp edge. $S_{\mathrm{COM}}$ is a Gaussian in the perpendicular distance from the object’s center of mass to the grasp axis; favoring small distances reduces gravitational torque about the grasp axis and bounds its worst case independently of the object pose. Finally, $S_{\mathrm{sep}}$ penalizes wrap contacts that are too close for the fingers to engage independently.

\begin{figure}[!t]
    \centering
    \includegraphics[width=1.0\linewidth]{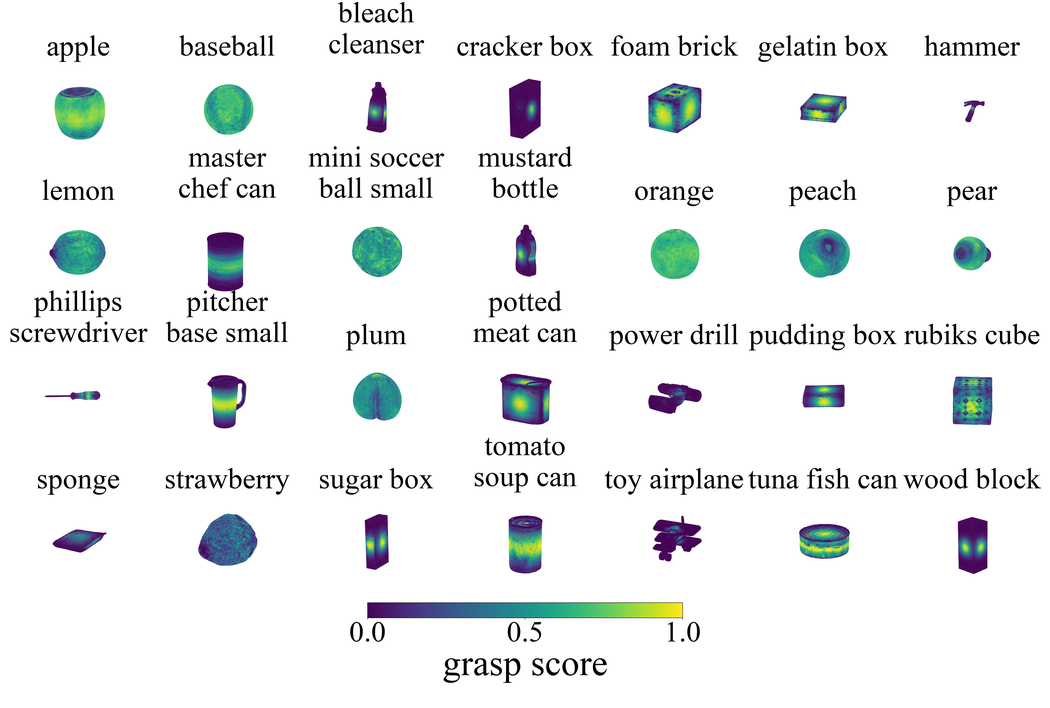}
    \caption{\textbf{Grasp-affordance heatmaps for 28 YCB objects.} Surface color denotes the normalized score $S_i$ assigned to a candidate thumb contact at each mesh vertex, using a common scale from low (purple) to high (yellow).}
    \label{fig:heatmaps}
    \vspace{-10pt}
\end{figure}

The resulting per-vertex scores form the heatmap, enabling constant-time lookup at runtime. The default heatmap favors grasps near the center of mass; Fig.~\ref{fig:grasping-pipeline} and Section~\ref{sec:steering} describe how it can be steered toward a specified object region. Fig.~\ref{fig:heatmaps} shows heatmaps for 28 YCB objects~\cite{calli2015ycb, calli2017ycb}. Objects such as the mini soccer ball, pitcher base, power drill, and toy airplane receive low scores, suggesting that their geometry is less compatible with the evaluated three-point antipodal pinch configurations.

\subsection{Steering the Grasp-Prior Heatmap}
\label{sec:steering}

Among the six terms in~\eqref{eq:affordance}, only $S_{\mathrm{COM}}$
imposes a spatial preference. To steer the grasp toward an arbitrary anchor
$\bm a$, we define its perpendicular distance to the $i$th grasp axis as
\begin{equation}
d_{\perp}(\bm a,i)
=
\left\lVert
\bm r_i - (\bm r_i^{\mathsf T}\bm g_i)\bm g_i
\right\rVert,
\qquad
\bm r_i = \bm a - \bm p_{t,i}.
\label{eq:anchor-dist}
\end{equation}
We then replace the center-of-mass term with an anchor-centered Gaussian:
\begin{equation}
S_i^{\bm a}
=
\frac{S_i}{S_{\mathrm{COM}}(i)}
\exp\!\left[
-\left(\frac{d_{\perp}(\bm a,i)}{\sigma_a}\right)^2
\right],
\label{eq:anchored-affordance}
\end{equation}
where $\sigma_a$ controls the spatial extent of the preferred grasp region.
The default heatmap is recovered by placing the anchor at the object's
center of mass, $\bm a = \bm c$, with $\sigma_a = \qty{30}{mm}$.

For language-guided steering, RexOmni~\cite{jiang2025detectpointprediction} maps an instruction to a 2D pixel, which is back-projected through the depth image to obtain a 3D anchor $\bm a$. Evaluating~\eqref{eq:anchored-affordance} then reweights all candidates in a few milliseconds without modifying their contact triples or any downstream module. Section~\ref{sec:extensions} demonstrates the effect of anchor-based steering on grasp selection.

\subsection{Online Hand-Pose-Adaptive Three-Point Target Selection}
\label{sec:interface}

Three-finger grasp synthesis has been studied through geometric, vision-based,
and force-closure formulations~\cite{park1992grasp,morales2006vision,ponce1995on}.
Our focus is complementary: we convert offline-scored candidate triples
from~\eqref{eq:affordance} into three-point targets online, adapting their
selection to the current object and hand poses.

At each control tick, the tracked object pose transforms all candidates into
the world frame. Each candidate is reduced to its thumb contact and
wrap-contact centroid, then ranked using the current fingertip positions and
palm orientation. The objective favors candidates whose thumb point and wrap
centroid are close to the current thumb fingertip and the mean index--middle
fingertip position, respectively, while also rewarding high cached affordance.
It rejects candidates that violate ground clearance and penalizes
palm--grasp-axis misalignment and mismatch between the candidate width and
current finger span.

The selected wrap centroid is split into two targets at nominal
\qty{20}{mm} offsets along $\bm u\times\bm g_i$, where $\bm g_i$ is the
selected candidate's grasp axis and $\bm u$ is an object-fixed reach direction
that follows the tracked object pose, yielding
\begin{equation}
\mathcal{I}
=
(\bm p_1,\bm p_2,\bm p_3)
\in \R^{3\times3}.
\label{eq:interface}
\end{equation}
Unlike the offline contacts, the resulting online targets are neither
reprojected onto the mesh nor refitted for curvature. The cached score
therefore approximates the quality of the thumb--centroid pair rather than the
realized three-point geometry.

\subsection{Last-Centimeter Policy for Grasp Adjustment and Lifting}
\label{sec:rlpolicy}

We adopt the model-based controllers of~\cite{lee2026hierarchical} to reactively reach the selected contact triples. However, naively squeezing at the target can be brittle: residual perception and convergence errors displace the achieved contacts from the intended ones, as shown in Section~\ref{sec:experiments}. We therefore train an RL policy to refine finger and arm motion over the final centimeter, while a scripted ramp supplies a small (\qty{1.0}{N}) contact force. The policy is trained once in simulation and deployed unchanged across objects, observing only finger joint states and its recent actions; the target contacts enter through the pre-grasp pose and the reach reward.

\subsubsection{Handoff Dataset}
\label{sec:handoff-dataset}

The handoff dataset defines the initial-state distribution for RL training. We generate 1,000 pre-grasp states across diverse object geometries using a sampling-based grasp-quality evaluator. For each candidate, inverse kinematics places the thumb \qty{25}{mm} and the index and middle fingers \qty{40}{mm} away from their respective target contacts. These distances are measured outward from the object along the corresponding mesh-surface normals returned by the heatmap generator. The resulting states approximate the hand configurations encountered when control is transferred to the learned policy while providing diverse initial conditions for training.

\subsubsection{Reward Structure}
Suitable three-contact configurations can provide force closure and resist external wrenches~\cite{nguyen1988constructing, li2003computing}. To encourage contact acquisition and subsequent lifting, we combine reach, contact, and lift rewards, with active terms determined by the episode stage, together with additional regularizing terms listed in Section~\ref{sec:impl-details}.

\paragraph{Reach Reward}
The reach reward encourages each fingertip twist to follow a guidance vector
field (GVF), adapted from~\cite{lee2026bcsdm}:
\begin{equation}
r_{\text{reach}}
=
\frac{1}{3}\sum_{f=1}^{3}
\exp\!\left(
-\frac{
\left\lVert \bm v_f-\mathrm{GVF}_f \right\rVert_w
}{
s\left(\left\lVert \mathrm{GVF}_f \right\rVert_w+\epsilon\right)
}
\right).
\end{equation}
Here, $\bm v_f$ is the current twist of fingertip $f$, and
$\mathrm{GVF}_f$ is the desired twist evaluated at its current pose. The
translational component guides the fingertip along a straight path from its
handoff pose to the target contact, with a reference speed that decays to zero
at the target. The rotational component aligns the fingertip pad with the
desired contact orientation through the $\mathrm{SO}(3)$ log map. Linear and
angular errors are combined using
\begin{equation}
\left\lVert \bm e \right\rVert_w^2
=
\left\lVert \bm e_{\mathrm{lin}} \right\rVert^2
+
w_{\mathrm{rot}}^2
\left\lVert \bm e_{\mathrm{ang}} \right\rVert^2,
\qquad
w_{\mathrm{rot}}=\qty{0.05}{m/rad},
\end{equation}
which converts angular error into linear-velocity units. The scale $s$
controls sensitivity to tracking error, while $\epsilon>0$ regularizes the
reward as the desired twist approaches zero.

\paragraph{Contact Reward}
To encourage all three fingers to engage the object, we use $r_{\text{contact}} = \sum_{f=1}^{3}\mathbb{I}(c_f)$, where $c_f$ indicates contact between finger $f$ and the target object. 
In simulation, $c_f$ is read from the simulator's contact solver, and the momentum observer (Appendix~\ref{sec:momentum-observer}) additionally decides when each finger has landed and when to transition to lifting. On hardware, where solver contacts are unavailable, the observer and tactile pads provide these signals.

\paragraph{Lift Reward}
The lift reward encourages upward palm motion at a desired speed while penalizing lateral motion:
\begin{equation}
r_{\text{lift}} =
\exp\!\left[-\left(\frac{v_z-v_{\text{des}}}{\sigma_z}\right)^2\right]
\exp\!\left[-\left(\frac{\lVert v_{xy}\rVert}{\sigma_{xy}}\right)^2\right].
\end{equation}
Palm velocity is computed from the arm's kinematic Jacobian. Here, $v_z$ and $v_{xy}$ are its vertical and lateral components in the world frame, $v_{\text{des}}$ is the desired upward speed, and $\sigma_z$ and $\sigma_{xy}$ set the respective reward widths.

\subsubsection{State Transitions}
\label{sec:state-transitions}
During contact acquisition, only $r_{\text{reach}}$ and $r_{\text{contact}}$ are active. Once all three fingers register contact, or after \qty{1.6}{s} with at least two contacts, the episode transitions to lifting: $r_{\text{lift}}$ replaces $r_{\text{reach}}$, while $r_{\text{contact}}$ remains active to encourage continued engagement.

\subsubsection{Implementation Details}
\label{sec:impl-details}
The policy outputs 15 actions at \qty{100}{Hz}: velocity residuals for the nine grasping finger joints, scaled by 5.0 and added to a model-based closing command, and velocity commands for six arm joints, scaled by \qty{0.15}{rad/s}; the lift motion itself is scripted. Its 66-dimensional observation stacks two consecutive steps of a 33-value block: the nine finger joint positions, the nine joint velocities, and the 15-dimensional previous action. The critic receives the same inputs. Actor and critic are multilayer perceptrons with exponential linear unit activations and two hidden layers of 128 units, trained with proximal policy optimization (PPO; $\gamma=0.99$, $\lambda=0.95$, no entropy bonus, and a fixed learning rate of $1.0\times10^{-3}$ during fine-tuning) using 1,024 environments and 64 steps per iteration. The reward uses $s=0.5$, $\epsilon=\qty{0.02}{m/s}$, $v_{\text{des}}=\qty{0.15}{m/s}$, and $\sigma_z=\sigma_{xy}=\qty{0.1}{m/s}$, with weights 3.0 (reach), 1.0 (contact), and 2.0 (lift), and two action-rate penalties ($-0.01$ and $-0.005$). The reported policy comes from a single training seed: a 460-iteration policy fine-tuned for 10 further iterations, selected as the best of 18 evaluated candidates trained with the same recipe.

\section{Experiments}
\label{sec:experiments}

Our experiments examine reaching accuracy, grasp-and-lift success, and robustness during closing and lifting. We begin with reach convergence to motivate a learned policy that operates from imperfect handoff configurations. We then compare against model-based squeezing and end-to-end RL, analyze handoff pose error and lift stability, and conclude with hardware demonstrations and spatial grasp steering.

\subsection{Reach Convergence Evaluation}
\label{sec:reach-test}

We first examine how accurately the reactive controller positions the fingertips before closing. We integrate its commanded joint velocities without physics simulation over 2,752 randomized cylinder poses, isolating reaching from contact dynamics. Convergence requires all fingertips to reach the specified tolerance while moving below \qty{0.025}{m/s}. The fine-adjustment timer starts when the largest fingertip error falls below \qty{140}{mm} and restarts if it exceeds that threshold.

At the deployed \qty{20}{mm} tolerance and \qty{3}{s} budget, convergence reaches $78.3\%$ (Table~\ref{tab:reach-tolerance-sweep}). Extending the budget to \qty{7}{s} yields only $80.0\%$, while accepting \qty{30}{mm} error at \qty{3}{s} yields $88.1\%$. Thus, additional reaching time offers limited improvement at the tighter tolerance. This motivates a closing policy that can handle residual positioning error, which we test below.

\begin{table}[!t]
    \centering
    \footnotesize
    \caption{Reach-convergence rate (\%) over 2,752 randomized poses. A reach converges when every fingertip is within the stated tolerance. $^\ddagger$Deployed configuration (\qty{20}{mm}, \qty{3.0}{s}).}
    \label{tab:reach-tolerance-sweep}
    \begin{tabular}{lrrrrrr}
        \toprule
        & \multicolumn{6}{c}{Fine-adjustment budget} \\
        \cmidrule(l){2-7}
        Tol. (mm) & 0.5\,s & 1\,s & 2\,s & 3\,s$^\ddagger$ & 5\,s & 7\,s \\
        \midrule
        10               &  0.0 &  0.0 & 32.6 & 69.6 & 77.7 & 78.9 \\
        15               & 13.1 & 30.2 & 61.5 & 75.5 & 79.1 & 79.7 \\
        20$^\ddagger$    & 44.2 & 50.8 & 68.4 & \textbf{78.3} & 79.6 & 80.0 \\
        30               & 53.1 & 75.4 & 84.9 & 88.1 & 90.3 & 91.5 \\
        50               & 84.8 & 99.2 & 99.3 & 99.3 & 99.3 & 99.3 \\
        \bottomrule
    \end{tabular}
\end{table}

At timeout, the controller hands off directly to closing if the largest fingertip error is at most \qty{50}{mm}. Otherwise, it reselects a grasp, with at most one retry.

\subsection{Experimental Settings and Baselines}
\label{sec:baselines}

Grasp-and-lift trials run in MuJoCo~\cite{todorov2012mujoco} at \qty{500}{Hz} physics and \qty{100}{Hz} control. Our method and the squeeze baseline share grasp selection and the initial-pose distribution, although trials are unpaired. Both solve reaching kinematically at reset and then have \qty{4}{s} to close, lift, and hold. The transition from reaching to closing is triggered when either the weighted midpoint of the three fingertips falls below a threshold or the fine-adjustment timeout is reached. Our closing policy is trained on 24 YCB objects, including all eight in Table~\ref{tab:per-object-success}.

Table~\ref{tab:per-object-success} reports fingertip accuracy at the reach-to-close handoff. Averaged over the three fingertips, converged reaches finish $13.3\pm2.1$\,mm from their targets, ranging from $9.8$ to $14.6$\,mm across objects. The residual is not spread evenly over the hand: the thumb converges to $8.7\pm3.5$\,mm while the two wrap fingertips stop at $14.2\pm5.0$\,mm and $17.1\pm3.7$\,mm. The controller gates its transition to closing on the weighted midpoint of the three fingertips rather than on the fingertips individually, and that quantity reaches $5.0\pm2.1$\,mm, below \qty{10}{mm} on every converged trial. The closing policy therefore takes over with the hand correctly centered on the object but at least one wrap fingertip still more than a centimeter from its intended contact, which is the residual the last-centimeter policy is trained to absorb.

The \emph{squeeze baseline} replaces the learned closing policy with fingertip guidance toward the selected contacts~\cite{lee2026bcsdm}, followed by a \qty{3.0}{N} grip command. This comparison assesses the contribution of learned closing and lifting.

\paragraph{End-to-end RL}
\label{sec:e2e-training}
This baseline learns reaching, closing, and lifting jointly using the closing policy's reward terms. It receives updated target contacts in the palm frame and is trained on 24 objects without object-identity input. Both policies receive approximately 31M training steps; the end-to-end baseline uses two seeds. At evaluation, it starts from the arm's home configuration with a \qty{9}{s} episode, including up to \qty{3}{s} for reaching. Its curriculum stalled before a forced transition to the full reach range, so its results reflect training difficulties as well as architectural differences.

\subsection{Overall Grasp-and-Lift Success}
\label{sec:overall-test}

All methods use the same success criterion: the object's origin must rise \qty{0.10}{m} above its reset height and remain there continuously for \qty{0.5}{s}, with less than \qty{0.03}{m} of palm-frame displacement since lift onset. 

We evaluate eight objects selected for the highest nominal success under our method, using a separate run after fixing the subset: potted meat can, master chef can, Rubik's cube, sugar box, apple, tomato soup can, foam brick, and a scaled-down pitcher base. Of the original 28 objects, four lack a pre-grasp at the deployed standoff, leaving 24 eligible objects. 

\begin{table}[!t]
    \centering
    \caption{Unweighted mean over eight selected objects. The $95\%$ intervals reflect per-object binomial variance. Ours and the squeeze baseline use 3,584 episodes per condition; end-to-end results pool two training seeds with at least 400 episodes per object each.}
    \label{tab:overall-success}
    \begin{tabular}{lc}
        \toprule
        Method & Nominal (\%)\\
        \midrule
        Ours          & $91.9\pm0.9$ \\
        Naive squeeze & $56.1\pm1.4$ \\
        End-to-end RL & $0.07$  \\
        \bottomrule
    \end{tabular}
\end{table}

\begin{table}[!t]
    \centering
    \caption{Per-object grasp-and-lift success. The final row is the unweighted mean over the eight objects reported in Table~\ref{tab:overall-success}. The end-to-end policy was not evaluated per object. The final column reports the fingertip-to-target distance at the reach-to-close handoff, averaged over the three fingertips and given as mean $\pm$ standard deviation, measured in closed-loop deployment runs.}
    \label{tab:per-object-success}
    \footnotesize
    \setlength{\tabcolsep}{3.2pt} 
    \begin{tabular}{lrrr}
        \toprule
        Object & Ours (\%) & Naive squeeze (\%) & Reach error (mm) \\
        \midrule
        Potted meat can & $100.0$ & $82.5$ & $13.8\pm1.3$ \\
        Master chef can & $99.1$ & $46.8$ & $21.1\pm2.0$\\
        Rubik's cube & $95.8$ & $86.6$ & $14.1\pm1.5$\\
        Sugar box & $93.9$ & $14.3$ & $13.8\pm2.5$\\
        Apple & $93.4$ & $70.1$ & $12.9\pm1.4$\\
        Tomato soup can & $87.5$ & $20.8$ & $13.2\pm0.9$\\
        Foam brick & $84.7$ & $71.7$ & $9.8\pm3.7$\\
        Pitcher base (scaled) & $80.5$ & $55.9$ & $14.6\pm0.8$\\
        \midrule
        Mean & $91.9$ & $56.1$ & $13.3\pm2.1$\\
        \bottomrule
    \end{tabular}
\end{table}

Our method reaches $91.9\%$ nominal success on the selected objects, compared with $56.1\%$ for squeezing (Table~\ref{tab:per-object-success}). 
The advantage holds on every one of the eight objects, from $+9.2$\,pp on the Rubik's cube to $+79.6$\,pp on the sugar box, so the aggregate gap is not an artifact of averaging over an uneven object mix. It is, however, concentrated: the sugar box and the tomato soup can together account for about half of it, and excluding them narrows the mean gap from $35.8$ to $23.3$\,pp.

The two controllers fail in different places. Our failures are dominated by episodes in which the object never reaches the lift threshold ($4.9\%$ of all trials), and they concentrate on the scaled pitcher base ($14.7\%$) and the foam brick ($11.9\%$); once an object is off the table it is seldom lost, with $2.7\%$ lifted but not held and $0.8\%$ slipping. The squeeze baseline fails both earlier and later: $23.8\%$ of its trials never lift, and a further $12.0\%$ lift without holding, with $8.3\%$ slipping --- ten times our slip rate. Its two collapse cases are the sugar box and the tomato soup can, where it lifts nothing in roughly half of trials ($49.2\%$ and $46.0\%$ never lifted) and slips in a fifth more ($23.8\%$ and $20.8\%$). These are the two objects on which our method reaches $93.9\%$ and $87.5\%$, so the aggregate difference is driven mainly by cases the squeeze baseline cannot handle at all rather than by a uniform margin.

Across all 24 eligible objects, nominal success is $40.3\%$ for our method and $29.8\%$ for squeezing. The larger gain on the selected subset shows that the benefit depends strongly on the object geometry.

\subsection{Sensitivity to Handoff Pose Error}

\begin{figure}[!t]
    \centering
    \includegraphics[width=1.0\linewidth]{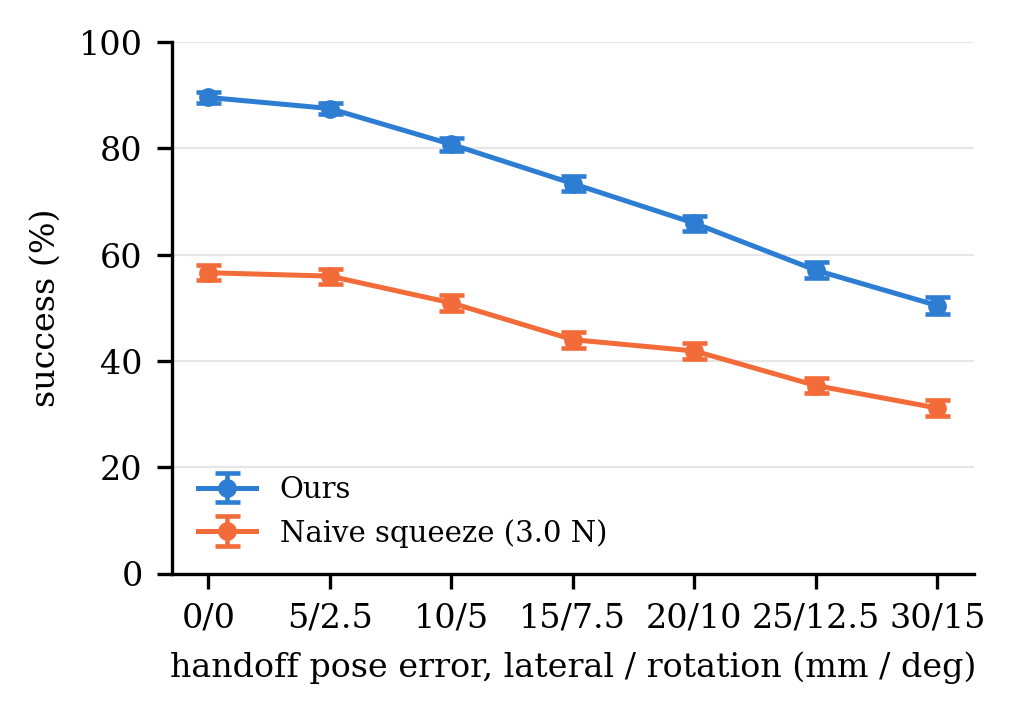}
    \caption{Grasp-and-lift success under handoff pose error. Each point is the unweighted mean across eight objects and 3,584 episodes, with $95\%$ bootstrap intervals.}
    \label{fig:handoff-pose-error}
\end{figure}

We test whether the learned closing policy tolerates errors at the transition from reaching. With the hand at its nominal handoff pose, we perturb the object's lateral position by up to $\pm e$ and its orientation by up to $\pm\theta$ about a random axis. Success decreases from $89.3\%$ without perturbation to $80.4\%$ at $e=\qty{10}{mm}$ and $\theta=\qty{5}{\degree}$ (Fig.~\ref{fig:handoff-pose-error}). The gradual decline supports tolerance to small handoff errors within this tested range.

\subsection{Contact Wrench and Lift Stability}
\label{sec:close-test}
\begin{figure}[!t]
    \centering
    \includegraphics[width=1.0\linewidth]{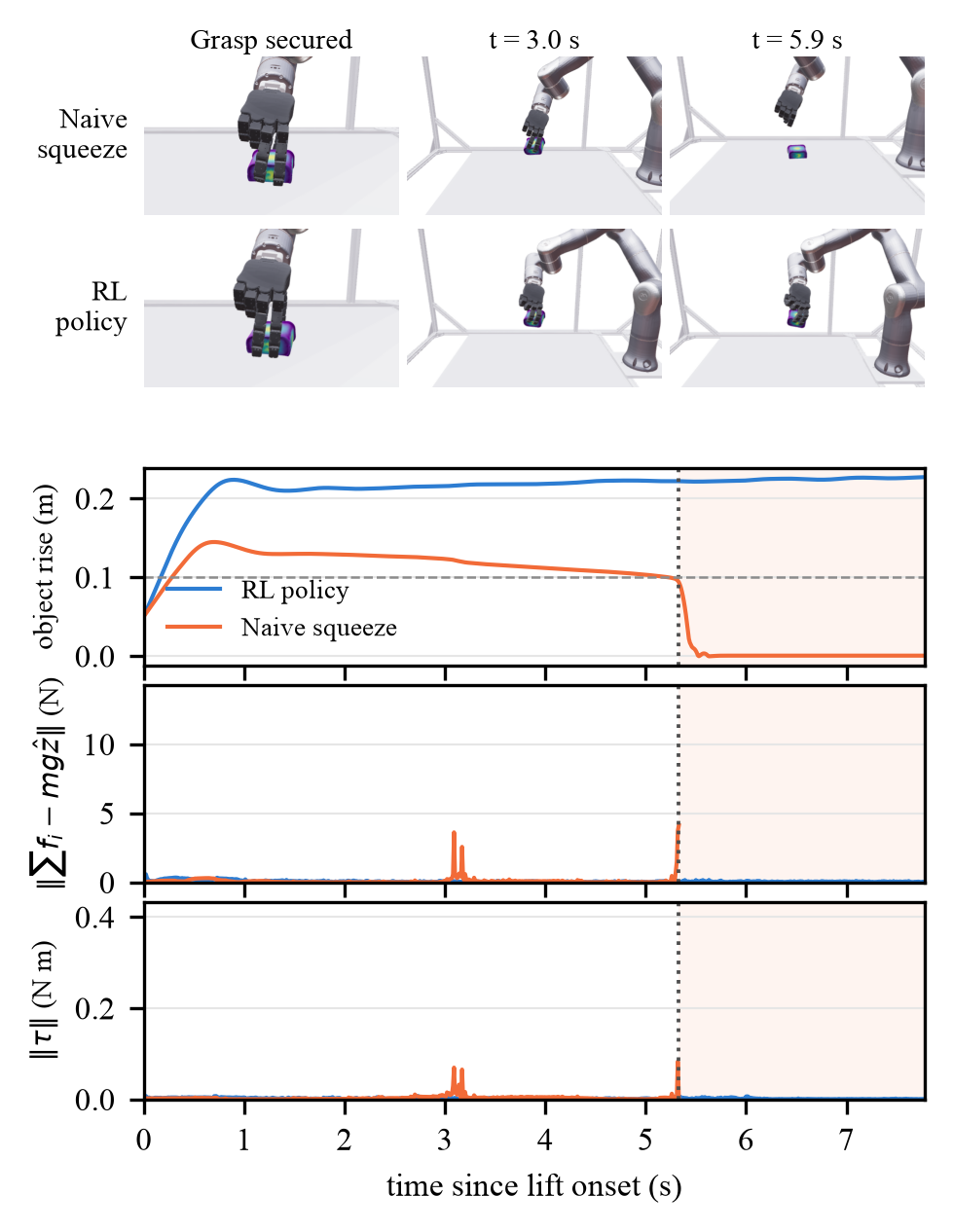}
    \caption[Close-and-lift on the potted meat can]{Representative potted-meat-can trial: squeeze (upper row) and learned policy (lower row). Plots show object rise and residual wrench; samples after squeeze contact loss are omitted. This deployment-controller simulation has no episode limit, unlike the \qty{4}{s} benchmark. Here, $t=0$ denotes a \qty{5}{cm} object rise.}
    \label{fig:wrench-comparison}
\end{figure}
In this trial, both controllers initially hold the object steady, with net force and torque close to zero. A steady wrench does not imply a secure grasp: around $t=5.3$\,s the squeeze baseline loses contact and the object slips free, while our policy maintains the grasp (Fig.~\ref{fig:wrench-comparison}).

Residual wrench alone does not distinguish grasp quality or internal squeezing forces: different contact-force distributions can produce the same net wrench during a static hold. The figure illustrates a particular failure and does not establish an aggregate advantage in force regulation.

\subsection{Hardware Demonstrations}

\begin{figure}[!t]
    \centering
    \includegraphics[width=1.0\linewidth]{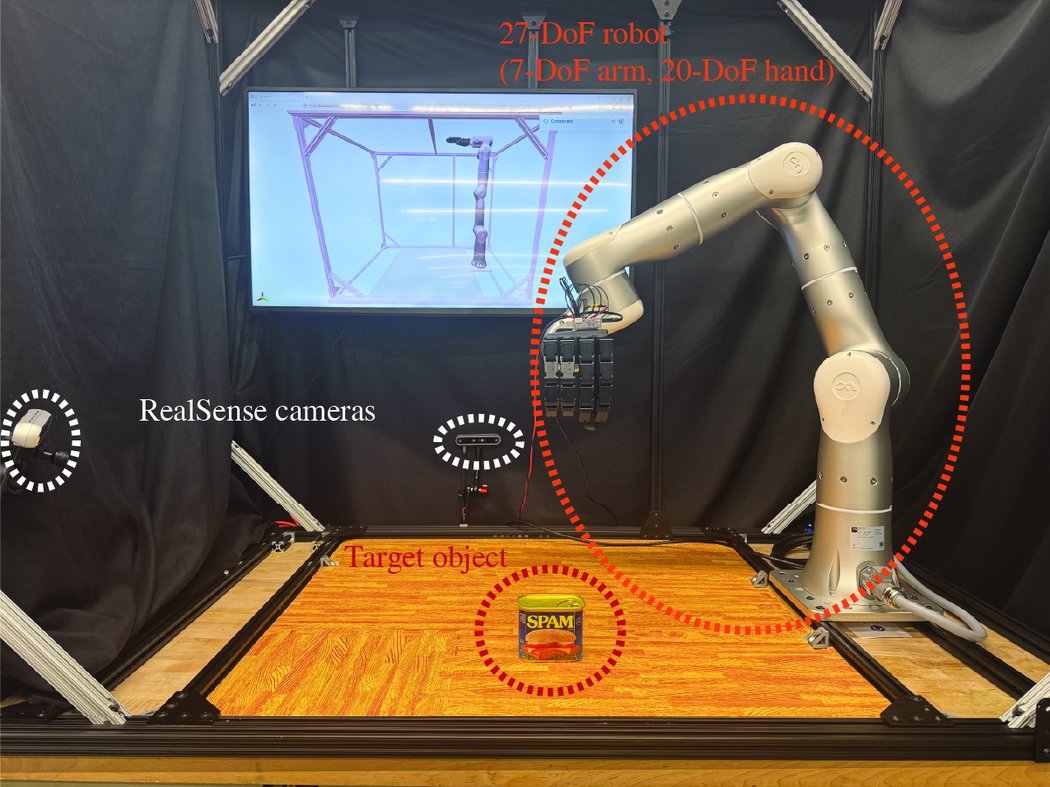}
    \caption{Hardware setup with a 7-DoF arm, a 20-DoF multifingered hand, and two external RealSense cameras. FoundationPose~\cite{wen2024foundation} tracks the target object.}
    \label{fig:hardware-setup}
\end{figure}

\begin{figure}[!t]
    \centering
    \includegraphics[width=1.0\linewidth]{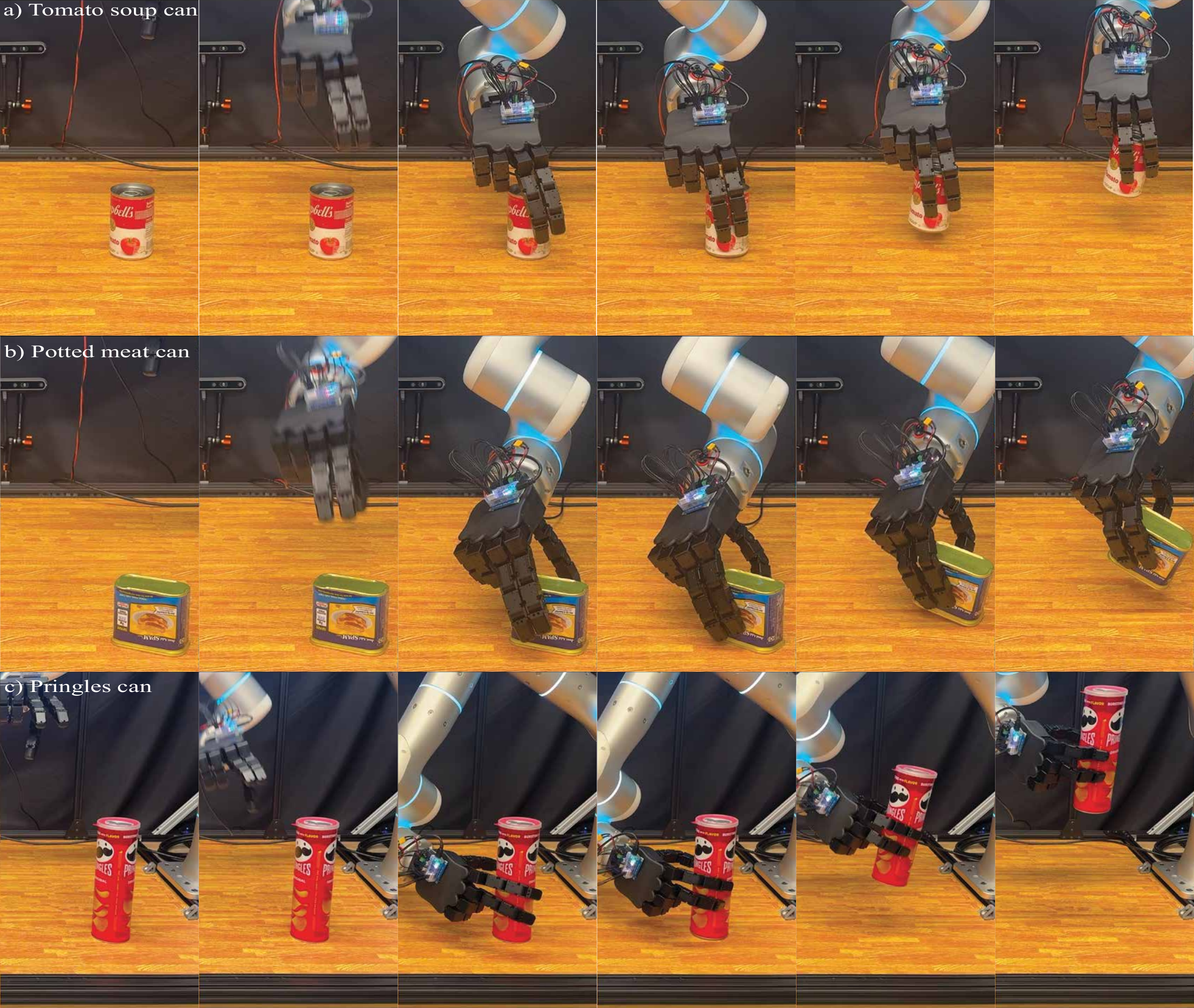}
    \caption{Hardware demonstrations, one time sequence per row: top-down grasps of (a) the tomato soup can and (b) the potted meat can, and (c) a side grasp of the unseen Pringles can.}
    \label{fig:hw-tests}
\end{figure}

We deploy the simulation-trained closing policy without fine-tuning on the 27-DoF platform (Fig.~\ref{fig:hardware-setup}). FoundationPose~\cite{wen2024foundation} supplies the object pose used to transform its precomputed heatmap for target selection. Reactive reaching approaches the selected contacts, with tactile sensing supplementing the momentum observer for contact detection. Fig.~\ref{fig:hw-tests} shows top-down grasps of two training objects and a side grasp of an unseen Pringles can. These demonstrations show execution of the integrated pipeline on hardware.

\subsection{Steering the Grasp}
\label{sec:extensions}

\begin{figure}[!t]
    \centering
    \includegraphics[width=1.0\linewidth]{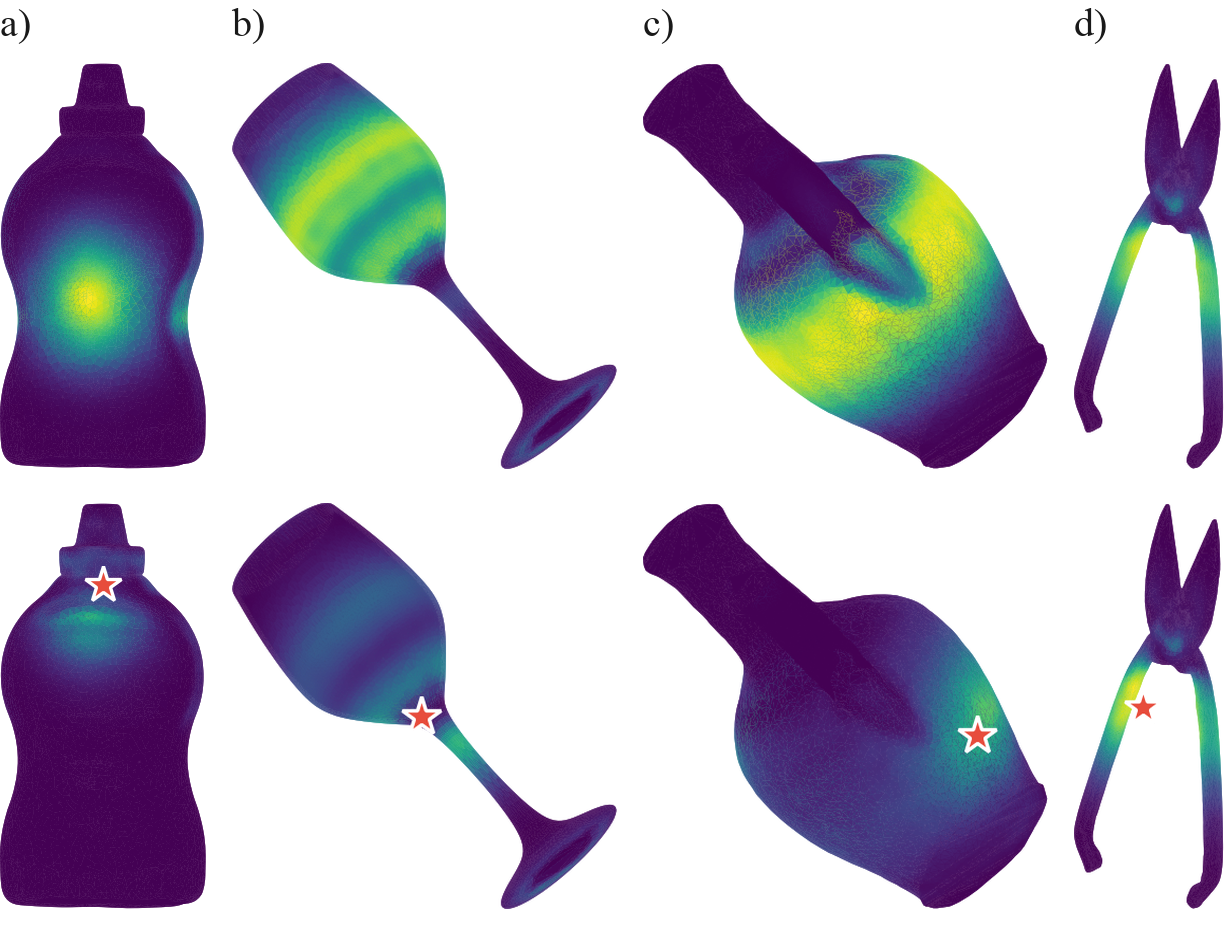}
    \caption{Heatmaps steered toward anchored positions on the object. Top row: the default heatmaps from object geometry alone. Bottom row: the same heatmaps after placing a 3-D anchor (red star) on the requested region of (a) the mustard bottle, top; (b) the wine glass, stem; (c) the pitcher, body; and (d) the wire cutters, handle. Each column shares one camera and one color scale, so the two rows are directly comparable.}
    \label{fig:language-steering-diverse}
\end{figure}

\begin{figure}[!t]
    \centering
    \includegraphics[width=1.0\linewidth]{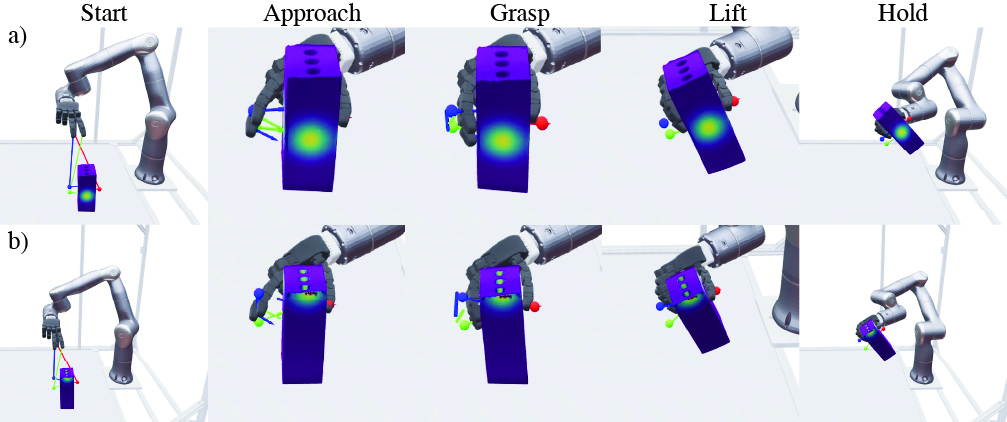}
    \caption{(a) Default grasp selection and (b) steering toward the top of the same object. Columns follow one execution from start to hold; the middle three overlay the heatmap, the sampled grasp positions, and the finger contact positions.}
    \label{fig:steer-trials}
\end{figure}

Finally, we test whether a spatial anchor shifts grasp selection toward a requested object region. Figs.~\ref{fig:language-steering-diverse} and~\ref{fig:steer-trials} show examples of the resulting heatmaps and grasps. In these experiments, we manually specify anchors at object parts or bounding-box faces.

\section{Limitations}
\label{sec:limitations}

The primary limitation of our system is its dependence on a complete object mesh for heatmap generation. Because our method is designed to grasp specific points on the object, it relies on accurate object pose tracking to align the heatmap and sample candidate grasp points. The success rate degrades steadily as handoff pose error grows, and our margin over the squeeze baseline narrows with it. Our system has been tested primarily on rigid objects with fully defined geometry, and fingertip pinch grasps do not suit every object: large or heavy objects can exceed either the hand's span (as the full-size pitcher base does) or the fingers' actuator limits. Future work could extend the method to deformable objects. Table~\ref{tab:overall-success} covers eight training objects selected by our own success rate; no held-out objects are evaluated in simulation; and the hardware results are qualitative. We expect difficulty to arise during reaching and closing rather than in contact selection: the hand may be unable to reach a well-scored triple without hitting the object, and a grasp that seats correctly may still exceed what the closing policy can hold. Our steering evaluation also measures the contact triple the controller commits to rather than the contacts it finally achieves.
\section{Conclusion} \label{sec:conclusion}

We presented a modular grasping system built around a pose-independent three-point interface that connects a model-based reaching controller to a learned RL policy for closing, lifting, and holding. Rather than training one policy to both reach and grasp, the framework applies model-based control to reaching, where object geometry and pose are available, and learning to the final centimeters of contact, which are hard to model. On a selected set of training objects, the learned closing policy outperforms a model-based squeeze controller, and hardware demonstrations show top-down and side grasps, including on an unseen object, without exposing the policy to the object's full geometry. Anchor-based steering moves the selected contacts toward a requested region. Future work includes deeper affordance- and language-conditioned interface selection, extension to $N$ fingers, and quantitative evaluation on held-out objects and on hardware.

\section*{Acknowledgment}
Anthropic's Claude was used to assist in implementing code. All generated code was verified, debugged, and validated experimentally by the authors.

\bibliographystyle{IEEEtran}
\bibliography{references}

\appendices
\section{Momentum Observer}
\label{sec:momentum-observer}
The momentum observer runs in both simulation and hardware, deciding when each finger has landed and when to transition from contact to lifting (Section~\ref{sec:state-transitions}).
We infer per-finger contact from external joint torques estimated by a generalized-momentum disturbance observer, which requires no acceleration measurements~\cite{deluca2003generalized, deluca2005sensorless}. Given
\begin{equation}
M(q)\ddot q + C(q,\dot q)\dot q + \tau_g(q)
= \tau + \tau_{\text{ext}},
\qquad \pi = M(q)\dot q,
\end{equation}
and using the identity $\dot M=C+C^\top$, which holds when $C$ is defined via Christoffel symbols, the observer estimate is
\begin{equation}
\begin{split}
\hat\tau_{\text{ext}}(t) = {} & K\Big[\pi(t)-\pi(0)\\
& - \int_0^t\big(\tau-\tau_g(q)+C(q,\dot q)^\top\dot q+\hat\tau_{\text{ext}}\big)\,dt'\Big],
\end{split}
\end{equation}
where all integrand quantities are evaluated at time $t'$, and $K$ is a positive-definite diagonal gain matrix. Under exact model dynamics, the estimate satisfies
\begin{equation}
\dot{\hat\tau}_{\text{ext}} = K(\tau_{\text{ext}}-\hat\tau_{\text{ext}}),
\end{equation}
and is therefore a first-order filtered estimate of the external joint torques, with bandwidth set by $K$.

The observer declares contact when a finger stalls under load:
\begin{equation}
c_f^{\text{obs}} =
\big(\max_j|\dot q_{f,j}|<\dot q_{\text{stall}}\big)
\;\wedge\;
\big(\max_j|\hat\tau_{\text{ext},f,j}|>\tau_{\text{thr}}\big),
\end{equation}
where $j$ indexes the joints of finger $f$, $\dot q_{\text{stall}}$ is the stall-velocity threshold, and $\tau_{\text{thr}}$ is the external-torque threshold. In simulation, $c_f$ combines the observer and solver contact flags; on hardware, it is the logical OR of the observer and tactile detections.

\end{document}